\documentclass[sigconf]{acmart}
\AtBeginDocument{%
  }

\renewcommand\footnotetextcopyrightpermission[1]{} %

\begin{document}
\fancyhf{} %
\pagestyle{empty} %

\title{MMCLIP: Cross-modal Attention Masked Modelling for Medical Language-Image Pre-Training}

\author{Biao Wu$^{1*}$\quad Yutong Xie$^{2*}$\quad Zeyu Zhang$^{3*}$\quad Vu Minh Hieu Phan$^4$\\
Qi Chen$^4$\quad Ling Chen$^1$\quad Qi Wu$^{4\dag}$\\
\vspace{0.2cm}
$^1$University of Technology Sydney\quad $^2$Mohamed bin Zayed University of Artificial Intelligence\\
$^3$The Australian National University\quad
$^4$University of Adelaide\\
\small $^{*}$Equal Contribution. $^{\dag}$Corresponding author: qi.wu01@adelaide.edu.au.}

\renewcommand{\shortauthors}{Trovato et al.}

\begin{abstract}
Vision-and-language pretraining (VLP) in the medical field utilizes contrastive learning on image-text pairs to achieve effective transfer across tasks. To further enhance visual and textual representation learning, many recent approaches adopt masked modeling strategies that randomly hide input tokens during training. Nevertheless, the adoption of masked modeling strategies in existing VLP methods gives rise to two key challenges in medical applications. First, current models struggle to accurately reconstruct key pathological features due to the scarcity of medical data. Second, most methods only adopt either paired image-text or image-only data, failing to exploit the combination of both paired and unpaired data. To this end, this paper proposes the \textbf{MMCLIP} (\textbf{M}asked \textbf{M}edical \textbf{C}ontrastive \textbf{L}anguage-\textbf{I}mage \textbf{P}re-Training) framework to enhance pathological learning and feature learning via unpaired data. \textit{First}, we introduce the attention-masked image modelling (AttMIM) and entity-driven masked language modelling module (EntMLM), which learns to reconstruct pathological visual and textual tokens via multi-modal feature interaction, thus improving medical-enhanced features. The AttMIM module masks a portion of the image features that are highly responsive to textual features. This allows MMCLIP to improve the reconstruction of highly similar image data in medicine efficiency. The EntMLM module identifies and masks key medical entities in the text with Named Entity Recognition (NER), and reconstructs them with support from visual features, enabling richer understanding of disease-related language. \textit{Second}, our MMCLIP capitalizes unpaired data to enhance multimodal learning by introducing disease-kind prompts. The experimental results show that MMCLIP achieves SOTA for zero-shot and fine-tuning classification performance on five datasets. 
 Our code will be available at \url{https://github.com/AIGeeksGroup/MMCLIP}. 
\end{abstract}




\maketitle

\section{Introduction}
\label{sec:introduction}
\begin{figure}[t]
\centering
\includegraphics[width=1.0\linewidth]{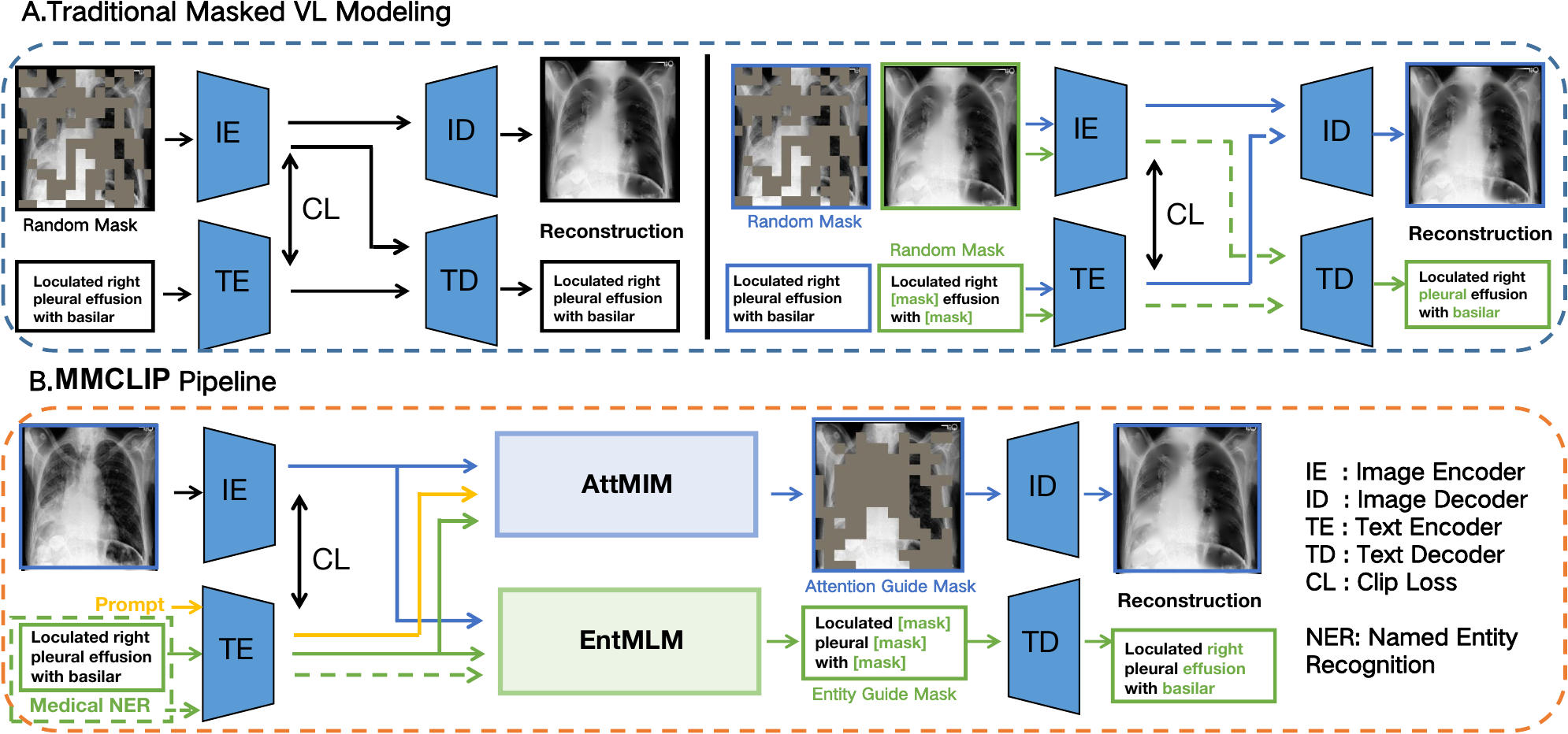} 
\caption{Modules A and B illustrate the distinctions between the existing VLP training frameworks and our proposed approach. Methods in module A only use the random mask strategy, and the reconstruction process does not involve multi-modal interaction. Module B shows that our method takes advantage of multi-modal interaction and uses an attention mechanism to guide the mask. }  
\label{fig:overall}  
\vspace{-3mm}
\end{figure}

Vision-and-language pretraining (VLP) has garnered increasing attention in the medical field, primarily due to its capability to transfer representations effectively across various downstream tasks, including zero- or few-shot recognition. This ability significantly reduces reliance on extensive annotated data and efficiently detects various pathologies. 
Most VLP approaches emphasize contrastive learning to understand the relationship between visual and linguistic elements 
\cite{american2014society}
Masked modelling has progressed and yielded promising results in pure visual \cite{he2022masked,bao2021beit,xie2022simmim}, and textual \cite{kenton2019bert} self-supervised representation learning.  Masked image modelling (MIM) reconstructs masked image segments using surrounding context, while masked language modelling (MLM) predicts masked words in sentences through adjacent context, both enhancing comprehension in their respective visual and linguistic domains. The success has inspired some researchers to explore the significance of MIM and MLM within the scope of VLP. This involves randomly masking elements in a single modality \cite{li2023scaling,sun2023eva} or across multiple modalities \cite{kwon2022masked,chen2023contrastive} during VLP, as shown in Figure~\ref{fig:overall} A.

Applying MIM and MLM to the medical domain still presents significant challenges, primarily due to two key limitations. 
\textit{First}, traditional random masking misaligns with the specific needs of medical data, which fails to capture pathological representations critical for clinical diagnosis. In MIM, random masking inadequately captures the subtle but critical variations in anatomical structures, crucial for precise medical diagnosis. Concurrently, in MLM, this approach can lead to a diminished focus on accurately predicting medical entities, generating less medically informative content. 
The latest research has started to explore attention-guided MIM \cite{wang2023hard,kakogeorgiou2022hide}, which are prone to finding crucial image regions for masking instead of random masking. However, constrained by the scarcity of medical data, models still struggle to accurately identify important medical areas during masking. 
\textit{Second}, MIM and MLM were initially conceived for single-modality self-supervised learning, not accounting for the strong interplay between medical images and clinical texts. This oversight limits their effectiveness in VLP scenarios where cross-modal interactions are vital. 
The advancement \cite{xie2023medim,chen2022multi} has proven that integrating modality interaction in MIM/MLM can enhance the recognition ability of meaningful regions, but heavily depending on the medical image-text pairs.

This motivates us to propose novel multimodal \textbf{M}asked \textbf{M}edical \textbf{C}ontrastive \textbf{L}anguage-\textbf{I}mage \textbf{P}re-Training, named \textbf{MMCLIP}. 
Our MMCLIP is designed to accomplish three self-supervised learning tasks, as shown in Figure~\ref{fig:overall} B. 
First, we design a novel attention-masked image modelling module (AttMIM) for medical images. It detects discriminative features to mask by harnessing feature interactions between vision and language modalities. 
Specifically, our AttMIM blends masks generated through image self-attention, image-report cross-attention, and prompt-driven attention. 
While the first two modules mine instance-level interactions, the prompt-driven attention exploits global-level interactions between images and common disease terms, independent of paired data. As such, even having missing corresponding reports, our prompt-driven attention still localizes pathological features by leveraging the global affinity with frequent diseases.
Secondly, we propose the entity-driven masked language modelling module (EntMLM) for medical reports, which masks informative entities and reconstructs the masked words. We also incorporate the learned image representations into the text decoder to aid the model in understanding medical entities with greater detail and nuance.
Besides, we employ standard contrastive learning to align medical images and reports.
To solve the inconsistent convergence rate problem between contrastive learning and masked modelling objectives, we first warm up the masked modelling tasks for specific iterations and then jointly train three tasks end-to-end.
 
Our MMCLIP is pre-trained on two large-scale medical datasets, one is MIMIC-CXR with paired chest X-ray images and reports, and the other is the PadChest with only chest X-ray images.
The learned image representations are transferred to five downstream classification datasets under zero-shot and fine-tuning settings, all of which achieved SOTA respectively.

Our contributions are three-fold: (1) we present a novel cross-modal attention masking, which is the first work to explore the potential of attention masking for MIM and MLM within a unified medical VLP framework, offering a new perspective to enhance the accuracy of medical data representation learning; (2) we develop a novel blending masking strategy that integrates attention-guided masking to capture discriminative pathological features, and common disease-prompt masking to enable unsupervised learning without relying on paired reports; (3) we conduct extensive experiments on medical image classification downstream tasks with improved zero-shot and fine-tuning performance. Our method surpasses strong competitors like MedKLIP, GLORIA, and CheXzero.

\vspace{-0.3cm}
\section{Related Works}

\subsection{Medical Vision Language Pretraining}

Driven by the effectiveness of self-supervised pre-training methods in NLP and CV, there is promising potential for customizing VLP methods to medical imaging analysis, such as lesion detection \cite{liu2019comparison} and segmentation \cite{wu2023bhsd,zhang2023thinthick,zhang2023segreg}. The most recent research exploring VLP is now centred around: improving feature extraction capabilities \cite{lu2019vilbert,li2019visualbert}; improving model structure \cite{li2020oscar,su2019vl}; improving training methods \cite{li2019visualbert,chen2020uniter}; improving model compatibility with different modalities \cite{xie2022unimiss}; introducing preprocessing modules \cite{mu2022slip}; introducing prior knowledge \cite{wu2023medklip}. As an application and extension of VLP in the medical field, Medical VLP focuses on understanding the content of medical images and texts. The latest outstanding research mainly focuses on the following three directions: 1) Model structures, particularly improvements in dual encoders \cite{jia2021scaling,yao2021filip} and fusion encoders \cite{li2019visualbert,kim2021vilt,tan2019lxmert,yu2021ernie}; 2) Scaling up training data, by collecting high-quality medical multi-modal datasets from various platforms to enhance representational learning \cite{huang2023leveraging,lin2023pmc,awais2023foundational}; 3) utilization of medical-specific prior knowledge to enhance model representational performance through prompts \cite{wu2023medklip,wang2022medclip}.

\subsection{Mask Image or Language Modelling}

The development of MIM in computer vision, represented by MAE~\cite{he2022masked}, SimMIM~\cite{xie2022simmim}, and iBOT~\cite{zhou2021ibot}, marks a significant advance in self-supervised learning. MAE and SimMIM innovate in image masking, while iBOT combines these advances with the pre-training technology of self-distillation. Recent approaches have extensively explored various fusions of SSL techniques. For example, CTITM \cite{chen2023contrastive} utilizes MIM and MLM to augment CLIP, thereby enhancing the expressive power of the visual coder. MaskVLM \cite{kwon2022masked} not only effectively integrates CLIP, MIM, and MLM, but also innovatively explores strategies for images and text to guide each other in generating masks, thus optimizing the synergy between modalities. However, these methods have limitations: they either don't make full use of the available multimodal information, depending instead on features from a single mode to perform tasks, or their exploration of masking strategies is confined to using random methods.

\subsection{Attention Masked modelling}

Traditional MIM pre-training methods typically require the model to predict the content of masked image patches using predefined masking strategies such as random masking, block-wise masking, and uniform masking. However, we believe that merely solving these tasks is insufficient; the model also needs to learn how to create challenging tasks.

In the realm of attention-based mask modeling, hard patch mining (HPM) stands out as a groundbreaking training paradigm for MIM, designed to enhance a model's comprehension of image content \cite{wang2023hard}. HPM method first has the model generate a challenging mask and then trains the model to predict the masked patches, similar to traditional methods. This approach enables the model to learn where it is worth applying masks while simultaneously learning how to solve the problems. Also, HPM introduces new metrics to measure the difficulty of the pre-training task by reconstructing the loss, which could help the model to focus more on challenging problems. 

Meanwhile, traditional random masking strategies, effective in general domains, fall short in medical contexts where precision is crucial~\cite{chen2022multi,huang2021gloria,muller2022radiological,zhang2022contrastive,wu2023medklip,zhou2022generalized}.  MedIM \cite{xie2023medim} mitigates the unique challenges posed by healthcare data by improving mask strategies.

However, current methods either rely solely on single-mode features for their attention mechanisms \cite{kwon2022masked} or have limited interaction between multimodal features, such as only employing feature matching \cite{xie2023medim}.
Our proposed method solves these problems well by employing cross-attention for multimodal feature interaction while performing attention-masked modelling on features that incorporate multimodal information, which allows the model to understand image and text information in a more integrated way.

\begin{figure*}[ht]    
    \centering
    \includegraphics[width=1\linewidth,scale=1]{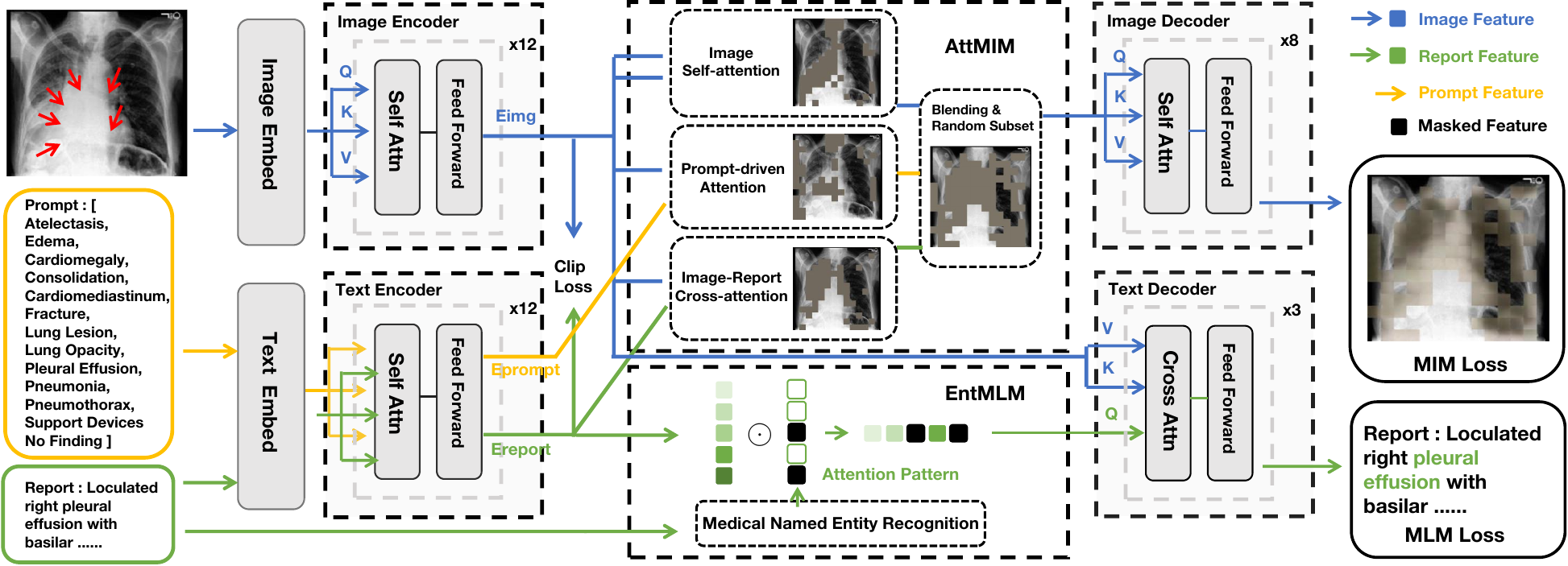}
    
    \caption{Our MMCLIP framework builds upon CLIP, integrating MIM and MLM modules, with a redesigned masking strategy to enhance model representation. The key contributions include three simple yet effective designs: generating masks through feature interaction, fusing masks to augment adaptability, and refining the text masking strategy with Medical Entity Recognition. Incorporating these designs into our multimodal pre-training framework significantly boosts the model's zero-shot performance.}
    \label{image_main}
\end{figure*}

\section{Methodology}

Figure \ref{image_main} illustrates the pipeline of the proposed MMCLIP. MMCLIP excels in representation learning by using image-report contrastive learning and masked modelling across both paired and unpaired medical image-report data. Central to MMCLIP are two innovative modules: Attention-Masked Image Modeling (AttMIM) and Entity-Driven Masked Language Modeling (EntMLM).

AttMIM capitalizes on vision-language interactions to selectively mask discriminative features in medical images. It seamlessly integrates image self-attention, cross-attention with medical reports, and prompt-driven attention. This multifaceted approach allows for the effective identification of pathological features, even when corresponding reports are unavailable.

EntMLM, on the other hand, concentrates on masking and reconstructing significant entities within medical reports, utilizing insights from image representations. This integration facilitates a more nuanced and detailed comprehension of medical terminology, enhancing the capabilities of both visual and textual learners within the MMCLIP framework.

\subsection{Image and Text Encoders}

\subsubsection{Image encoder}

We employ the Vision Transformer (ViT)~\cite{dosovitskiy2020image} with a patch size of 16 as the image encoder, initialized with the weights of the clip-base, to process the given input medical images $x$. 
Initially, $x$ is transformed into a sequence of flattened patches. These patches are subsequently embedded and introduced into a 12-layer transformer followed by a linear projection layer, denoted as $F(\cdot)$ and $Linear(\cdot)$, resulting in encoded representations of visual tokens: 

\vspace{-2mm}
\begin{equation}
E_{\text{img}} = Linear(F(x)) \in \mathbb{R}^{N_{\text{img}} \times C},
\end{equation} where $C$ is the encoding dimension and $N_{img}$ denotes the number of patches.

\subsubsection{Text encoder}
The text encoder also consists of a 12-layer transformer and a linear projection layer, initialized with the weights of the clip-base, denoted as $T(\cdot)$ and $Linear(\cdot)$. 
Our method involves the text \( t_{report} \) and \( t_{prompt} \) being tokenized using Byte Pair Encoding (BPE) into \( N_{\text{report}} \) and \( N_{\text{prompt}} \) tokens, which are then transformed into embeddings. These embeddings serve as the input for the text encoder. The encoder \( T(\cdot) \) processes these to create subword features:
 
\begin{equation}
E_{\text{report}} = Linear(T(t_{report})) \in \mathbb{R}^{N_{\text{report}} \times C},
\end{equation}
\begin{equation}
E_{\text{prompt}} = Linear(T(t_{prompt})) \in \mathbb{R}^{N_{\text{prompt}} \times C},
\end{equation}

where \( C \) represents the feature dimension.

\begin{figure*}[h]
\centering
\resizebox{1\linewidth}{!}
{\includegraphics[width=1\linewidth]{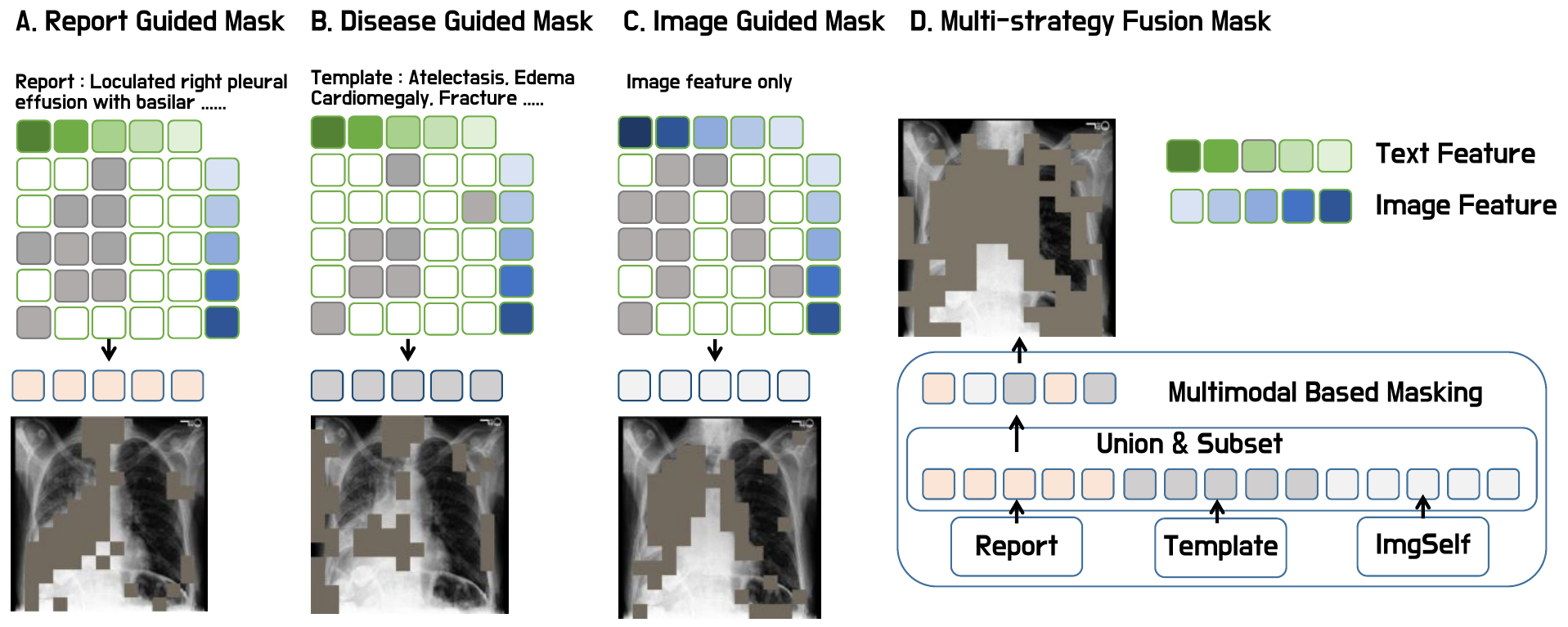} }
\vspace{-2mm}
\caption{The working mechanism of function M: The features obtained from different text data correspond to image features through the Cross Attention layer, querying the corresponding activated features to identify different regions of interest. These regions of interest are then combined, and a 25\% subset is randomly selected to obtain the final Mask result. Unlike MediM, which only uses dot products to activate cross-modal response features, MMCLIP employs cross-attention for deeper feature activation. This allows the model to induce more precise key areas from text features.}  
\label{fig:image_8}  
\vspace{-2mm}
\end{figure*}

\subsection{Attention-masked Image Modeling Module}

To address the limitations of random masking in medical imaging, we propose an attention-masked image modelling (AttMIM) module. It is divided into two phases: in the attention extraction phase, it uses an attention mechanism to focus on features that are relevant to the medical reports, disease prompts, and the image features themselves; in the attention-based mask generation and blending phase, it merges highly activated regions of these features to generate corresponding masks that accurately mask key pathology features (see Figure~\ref{fig:image_1}).

\subsubsection{Attentions extraction}

\textit{Image-report cross-attention} effectively identifies image regions that are relevant to the diagnostic report. The detailed information contained within these reports directs the model's focus towards the most crucial pathological features. Given the diversity of human organs, their unique physiological structures, and the specific pathological details associated with each, diagnostic reports typically emphasize areas presenting abnormalities. This guidance enables the model to prioritize regions abundant in diagnostic information, ensuring a more focused and informed analysis of medical images.

Based on the cross-attention mechanism, the image features after interacting with the report features can be obtained as follows:
\begin{align}
A_r = \text{softmax}\left(\frac{E_{\text{img}} E_{\text{report}}^T}{\sqrt{d_k}}\right)\cdot E_{\text{report}}, 
\end{align}
where $d_{k}$ is the dimension of $E_{\text{report}}$. 

\begin{figure*}[t]
\centering
\includegraphics[width=1\linewidth]{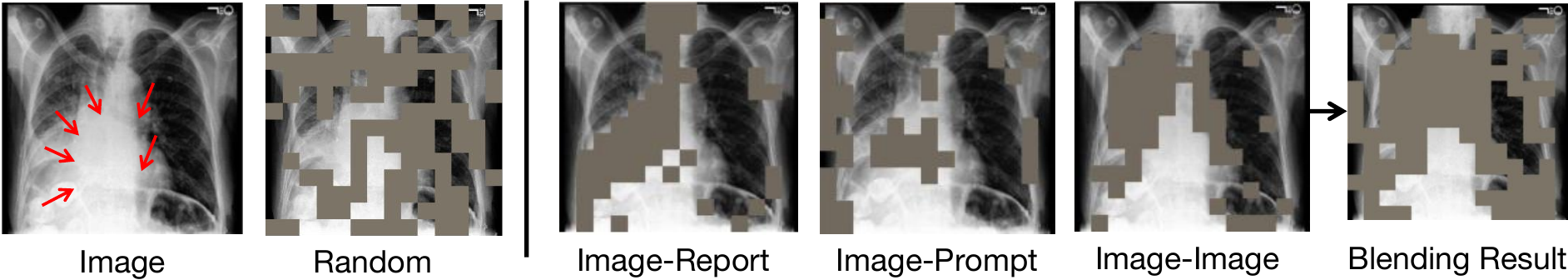} 
\caption{Different masking strategies result in varying concerns regarding mask application. Masks that are specifically tailored to the pathological characteristics of the lesion area are more effective than those applied randomly.}  
\label{fig:image_1}  
\vspace{-3mm}
\end{figure*}

\textit{Prompt-driven attention} can effectively recognize the image regions related to the disease prompts denoted as in Figure \ref{fig:image_1}. 
Meanwhile, as shown in Figure \ref{image_main}, the disease prompt contains 14 common diseases and medical devices under the chest organ, and this rich variety of diseases helps the model to focus on all potential lesions, thus improving the compatibility of MIM with all candidate regions where lesions may be present.

Besides, including disease prompts offers a significant advantage in scenarios where only images are available and the corresponding diagnostic reports are missing, acting as a bridge to fill the information gap. By supplementing image data, these prompts guide the model in identifying relevant pathological features, enabling effective representation learning and a deeper understanding of medical images without needing textual descriptions.
Based on the cross-attention mechanism, the image features after interacting with the disease prompt features can be obtained as follows:
\begin{align}
A_p = \text{softmax}\left(\frac{E_{\text{img}} E_{\text{prompt}}^T}{\sqrt{d_k}}\right)\cdot E_{\text{prompt}} , 
\end{align}
where $d_{k}$ is the dimension of $E_{\text{prompt}}$.

\textit{Image self-attention} directs the model to focus on the most challenging parts of the image features during convergence, which tend to be more difficult regions \cite{wang2023hard}, thus improving the model's ability and efficiency in understanding the image \cite{zhang2024jointvit, ji2024sine}. Based on the self-attention mechanism, the image features after interacting with the image features themselves can be obtained as follows:
\begin{align}
A_i = \text{softmax}\left(\frac{E_{\text{img}} E_{\text{img}}^T}{\sqrt{d_k}}\right)\cdot E_{\text{img}} , 
\end{align}
where $d_{k}$ is the dimension of $E_{\text{img}}$. 

\subsubsection{Attention-based mask generation and blending.}
We introduce the function $ M(A, \lambda) $, as shown in Fig. \ref{fig:image_8}, which retrieves indices corresponding to the top values in feature matrix $A$, with the number of indices determined as a proportion $ \lambda_1 $ of the matrix’s size. This function effectively transforms these indexes into a mask corresponding to the input matrix, which is the high-activated attention masking results in the attention extraction. 
We blend the masks produced by different strategies in a union manner and define this result as \( M_b \), formulated as:
\begin{align}
M_b = M(Ar; \lambda_1) \cup M(Ap; \lambda_1) \cup M(Ai; \lambda_1).
\end{align}
Subsequently, a subset is randomly selected in $ \lambda_2 $ proportion to constitute the final outcome of AttMIM masking process. We defined it as \( \mathbb{M}_{\text{AttMIM}}\):
\begin{align}
\mathbb{M}_{\text{AttMIM}}= {\rm RandomSubSet}(M_b, \lambda_2).
\end{align}
Finally, we apply the mask $\mathbb{M}_{\text{AttMIM}}$ to \( E_{\text{img}} \), rendering it as the feature input for MIM, which can be expressed as \begin{align}
{\rm MIM}_{\text{input}}= \mathbb{M}_{\text{AttMIM}} \cdot E_{\text{img}}.
\end{align}
The comparison results in Figure \ref{fig:image_1} show that the interactions of image self-attention, image-report cross-attention, and prompt-driven attention focus on different lesion areas to varying degrees. By integrating the strengths of these, AttMIM obtains a final mask that balances specificity and comprehensiveness.

\subsection{Entity-driven Masked Language Modeling Module}

Unlike natural language where fluency is emphasized, medical texts prioritize the accuracy of medical entity terms over textual smoothness. The typical autoregressive models in general MLM tasks may not be essential for medical MLM. Thus, we design an EntMLM module to target disease entities for masking. 

\begin{figure*}[t]
\centering
\includegraphics[width=1\linewidth]{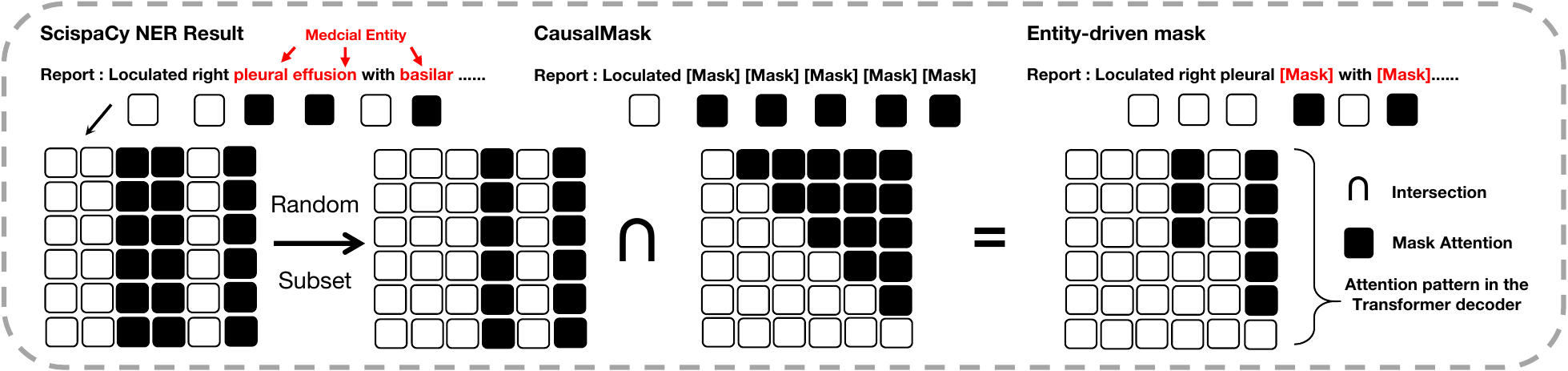} 

\vspace{-2mm} 
\caption{Illustration of entity-driven mask generation, which combines the CausalMask with the mask guided by the medical NER to generate the final result.}  
\label{fig:image_4}  
\end{figure*}

For EntMLM, we adopt ScispaCy  a specialized Named Entity Recognition (NER) tool for biomedical and scientific text analysis. It excels in identifying medically relevant entities, such as diseases, drugs, medical terms, treatments, and symptoms, from healthcare-related documents like medical records and research papers. As shown in Figure~\ref{fig:image_4}, we first identify medical entities in the report using ScispaCy, then retrieve their position indexes, and use these indexes to guide the mask generation. The resulting mask can be denoted as \( M_e \).

Notably, this mask is not directly applied to the original report text but is instead integrated with an upper triangular mask in the text decoder, which we defined as $\text{CausalMask}$, where their intersection is taken. CausalMask plays a key role in self-supervised learning tasks for natural language processing, especially in generative tasks. It facilitates the learning of the structure of the language and the generation of coherent text by ensuring that the model's predictions rely only on past and present information. The implementation of this approach enhances the ability of the model to process and generate natural language more efficiently. Hence, the mask result of  EntMLM, we defined it as \(\mathbb{M}_{\text{EntMLM}}\): 
\begin{align}
\mathbb{M}_{\text{EntMLM}} = \text{RandomSubSet}(M_e, \lambda_3)  \cap \text{CausalMask}.
\end{align}
As shown in Figure~\ref{fig:image_4}, this approach allows the model to temper its focus on text fluency, thereby enhancing its integration with image features and improving its understanding of medical semantic entities. Finally, we apply the mask $\mathbb{M}_{\text{EntMLM}}$ to \( E_{\text{report}} \), rendering it as the feature input for MLM, formulated as:

\begin{align}
\text{MLM}_{\text{input}}= \mathbb{M}_{\text{EntMLM}} \cdot E_{\text{report}}.
\end{align}

\vspace{-5mm}
\subsection{Objective Functions}

\subsubsection{Objective function for image-report alignment.} 

Inspired by \cite{li2021align}, our approach focuses on aligning image and text features prior to their fusion by the multimodal decoder. This pre-alignment strategy simplifies the process of cross-modal learning for the multimodal decoder, facilitating a more effective integration of visual and textual information. We follow \cite{radford2021learning} and compute the objective alignment function $ Loss_{\text{align}} $ by exploiting the ﬁne-grained correspondences between $ E_{\text{img}} $ and $ E_{\text{report}} $, formulated as:

\begin{equation*}
    \begin{aligned}
        \text{Loss}_{\text{align}} = -\frac{1}{N} \Bigg( &
        \underbrace{\sum_{i=1}^N\log{\frac{\exp(x_i^\top y_i / \sigma)}{\sum_{j=1}^{N} \exp(x_i^\top y_j / \sigma)}}}_\text{image-to-text} \\
        & + \underbrace{\sum_{i=1}^N\log{\frac{\exp(y_i^\top x_i / \sigma)}{\sum_{j=1}^{N} \exp(y_i^\top x_j / \sigma)}}}_\text{text-to-image} 
        \Bigg)
    \end{aligned}
\end{equation*}

where $ x_i $  and $ y_j $ are the $ E_{\text{img}} $  in the $i$-th pair and the $ E_{\text{report}} $  in the  $j$-th pair, respectively. $N$ is the batch size, and \(\sigma\) is the temperature to scale the logits.

\subsubsection{Objective function for image reconstruction.} 
The image decoder, inspired by \cite{he2022masked}, consists of an 8-layer transformer, which is essential for pre-training image reconstruction. This component integrates encoded visible patches and mask tokens to reconstruct the image. We defined this decoder as $D_{\text{MIM}}(\cdot)$.
Reconstruction accuracy is quantified by the Mean Squared Error loss computed solely on masked patches, shown as: \begin{align}
Loss_{\text{MIM}} = \left \| y^{\text{MIM}} , x\right \|^2,~\text{where}~y^{\text{MIM}} =  D_{\text{MIM}}( \text{MIM}_{\text{input}}).
\end{align}

\subsubsection{Objective function for report reconstruction.} 

The text decoder in our model, following \cite{yu2022coca}, features a 3-layer transformer. 
The multi-modal decoder combines visual and textual data, enhancing the representations. It uses a combination of causal and disease-guided masks and integrates visual encoder outputs via alignment from contrastive learning and the cross-attention mechanism.

Moreover, when introducing multi-modal information, the model can leverage visual information to predict masked textual entities, significantly aiding the model's understanding during the pre-training phase.
This allows the decoder to predict text while simultaneously incorporating image context. Such an approach guarantees efficient and flexible integration of diverse modalities in multi-modal learning tasks. Meanwhile, the training of text generation is to maximize the conditional likelihood of the paired text \(y\) under the forward autoregressive factorization:

\begin{equation}
Loss_{\text{MLM}} = - \sum_{t=1}^{T} \log P_{\theta}(y_{t}| y_{<t}, x),~\text{where}~y_{t} \in \text{MLM}_{\text{input}}.
\end{equation}

\subsubsection{Final objective function.} 

In the medical field, the scenario of unpaired data is common and significant, mainly due to data access limitations, ethical and privacy considerations, and the diversity and complexity of medical data. These factors make it difficult to obtain perfectly paired medical data in practical applications and also highlight the need to develop advanced algorithms capable of handling such data. MMCLIP Integrating prompt-driven attention and image self-attention facilitates the generation of masks for image data, which lacks corresponding reports. 
This approach improves the training process, supports compatibility with unpaired image data, and ensures that even data solely involved in the MIM task contributes to enhanced model representation learning. 
Our whole training process mixes paired data and unpaired data, thus the final objective function consists of these two parts as well.

As shown in Figure~\ref{image_main}, the paired data effectively contributes to the MIM task by the image-report cross-attention, prompt-driven attention, and image self-attention together. Besides, it aids in the MLM and alignment tasks through the combined use of image and report representations, demonstrating the advantages of multimodal feature interaction. The objective function can be defined as:

\begin{align}
Loss_{\text{pair}} =  Loss_{\text{align}} + Loss_{\text{MIM}} + Loss_{\text{MLM}}.
\end{align}
Meanwhile, the unpaired data is unable to complete the MLM and alignment tasks due to the lack of corresponding reports, but can effectively participate in the MIM task with the help of prompt-driven attention and image self-attention.
 
The objective function can be formulated as:
 
\begin{align}
Loss_{\text{unpair}} =  Loss_{\text{MIM}}.
\end{align}
The final objective is the combination of $ Loss_{\text{pair}} $ and $ Loss_{\text{unpair}} $ as:
\begin{align}
Loss_{\text{final}} = Loss_{\text{pair}} + Loss_{\text{unpair}}.
\end{align}

\section{Experiments}

\subsection{Dataset}
The experimental data is divided into pre-training data and downstream task data. The pre-training data includes MIMIC \cite{johnson2019mimic} data that can complete multi-modal tasks and Padchest \cite{bustos2020padchest} data used as unpaired image-only data. The downstream task data includes binary classification datasets CovidX \cite{pavlova2022covid} and Pneumonia \cite{rsna-pneumonia-detection-challenge}, as well as multi-label classification of 14 types of diseases datasets CheXpert \cite{irvin2019chexpert} and Xray14 
\cite{wang2017chestx}. 
Please refer to \textit{Appendix} for details.

\subsection{Implementation Details}

We initialize the image and text encoders with official clip pre-training weights.
The experiments are conducted using a single NVIDIA A100. SGD is employed as the network optimizer with a momentum of 0.9 and an initial learning rate of 5e-5. The input resolution is set to 224 × 224, and the batch size is 64. 
For a warm-up, we first optimize the masked modeling objectives for 15,000 iterations and then jointly train all objectives end-to-end. 
After pre-training, we keep only the image and text encoders for downstream tasks. Using the CLIP \cite{radford2021learning} approach, zero-shot classification is done by matching the image and category name features from both encoders. Meanwhile, the image encoder can be used independently for fine-tuning classification tasks.
About evaluation metrics, following previous works~\cite{tiu2022expert,xie2023medim}, we adopt the area under the ROC curve (AUC), ACC, and F1 scores, which are standard metrics to evaluate classification tasks.

\begin{table*}[t]
\centering
\caption{Zero-shot performance of various pre-training models on three datasets.}
\resizebox{0.8\linewidth}{!}
{
\begin{tabular}{l ccc ccc ccc ccc}
\hline
\textbf{Models}  & \multicolumn{3}{c }{\textbf{CheXpert}} & \multicolumn{3}{c }{\textbf{PadChest-Seen 8}} & \multicolumn{3}{c}{\textbf{Pneumonia}}  \\
 & \textbf{AUC} & \textbf{F1} & \textbf{ACC} & \textbf{AUC} & \textbf{F1} & \textbf{ACC} & \textbf{AUC} & \textbf{F1} & \textbf{ACC}   \\
\hline
ConVIRT \cite{zhang2022contrastive} & 52.10 & 35.61 & 57.43 & 74.31 & 23.58 & 80.12 & 79.21 & 55.67 & 75.08   \\
GLORIA \cite{huang2021gloria} & 54.84 & 37.86 & 60.70 & 74.56 & 24.02 & 80.75 & 70.37 & 48.19 & 70.54   \\
BioViL \cite{bannur2023learning} & 60.01 & 42.10 & 66.13 & 71.17 & 20.75 & 79.58 & 84.12 & 54.49 & 74.43  \\
BioViL-T \cite{bannur2023learning} & 70.93 & 47.21 & 69.96 & 76.17 & 27.41 & 85.32 & 86.03 & 62.56 & 80.04  \\
CheXzero \cite{tiu2022expert} & 87.90 & 61.90 & 81.17 & 76.75 & 28.68 & 87.37 & 85.13 & 61.49 & 78.34   \\
MedKlip \cite{wu2023medklip}  & 89.97 & 63.67 & 84.32 & 84.97 & 35.17 & 90.68 & 86.94 & 63.42 & 80.02 \\
\hline
MMCLIP  & 90.53 &67.05 & 86.04 &85.79 &33.50 &91.19 &86.16 &62.57 &80.46 \\
+ Ensemble & \textbf{91.45} &\textbf{69.18}	& \textbf{86.28} &\textbf{86.31} &\textbf{35.61} &\textbf{92.16} &\textbf{86.95}	& {63.30} &\textbf{80.51} \\
\hline
\end{tabular}
}
\label{zeroshot}

\end{table*}

\section{Results}

\subsection{Zero-shot Classification}
We compare our MMCLIP with different pre-training methods, including ConVIRT~\cite{zhang2022contrastive}, GLORIA \cite{huang2021gloria}, BioViL \cite{bannur2023learning}, BioViL-T\cite{bannur2023learning}, CheXzero \cite{tiu2022expert}, and MedKlip \cite{wu2023medklip}, for zero-shot classification. 
The results in Table \ref{zeroshot} and Table \ref{tab:padchest} show that our MMCLIP exhibits superior zero-shot performance across all these scenarios on three datasets, even for the unseen class recognition on the PadChest dataset.
This may be because it is difficult for existing models to understand the feature information contained in the whole image and report, while the proposed AttMIM and EntMLM can utilize the rich multimodal information of the medical data itself to help our MMCLIP to understand the more critical medical knowledge, and even understand the complex medical features that have not been mentioned in the report of the training set.

Specifically, in Table \ref{zeroshot}, compared to the second-best MedKlip, the AUC scores of our MMCLIP have increased as follows: from 89.97 to 90.53 on the CheXpert dataset, from 84.97 to 85.79 on the PadChest-seen dataset, and from 85.94 to ours 86.16 on the RSNA Pneumonia dataset. Meanwhile, we used the ensemble learning method to combine the weights obtained from 6 different groups of parameters, which can improve the performance further on most datasets.
In Table \ref{tab:padchest}, we comparatively analyze the zero-shot performance of MMCLIP on the PadChest dataset for unseen classes. Only 8 categories in PadChest strictly match the visible categories in the MIMIC data, so we only report the test results of 8 visible categories in Table \ref{zeroshot}. There are 12 categories that either cannot be strictly matched, but cannot be counted as unseen classes, or do not appear in the test set, so we report at most 173 unseen classes.  Our analysis of the 10 and 20 most numerous of these unseen classes reveals that the zero-shot performance of MMCLIP has a 4\% to 5\% lift.

\begin{table}[h]
\centering
\caption{Comparison of zero-shot AUC scores on the PadChest dataset. Unseen 10 represents the 10 categories with the highest number of unseen categories.}

\resizebox{0.85\linewidth}{!}
{
\begin{tabular}{ l c c c c }
\hline
\textbf{PadChest}  & \textbf{Unseen 10 } & \textbf{Unseen 20 } & \textbf{Unseen 173} \\ \hline
CheXzero \cite{tiu2022expert}   & 59.75 & 63.11  & 65.76  \\  
MedKlip \cite{wu2023medklip}    & 60.07 & 53.73  & 64.16   \\ 
MMCLIP   & \textbf{65.09} & \textbf{67.14}  & \textbf{70.00}    \\ \hline
\end{tabular}
}
\label{tab:padchest}

\end{table}

\begin{figure}[h]
\centering
\resizebox{1\linewidth}{!}
{
\includegraphics[width=1\linewidth]{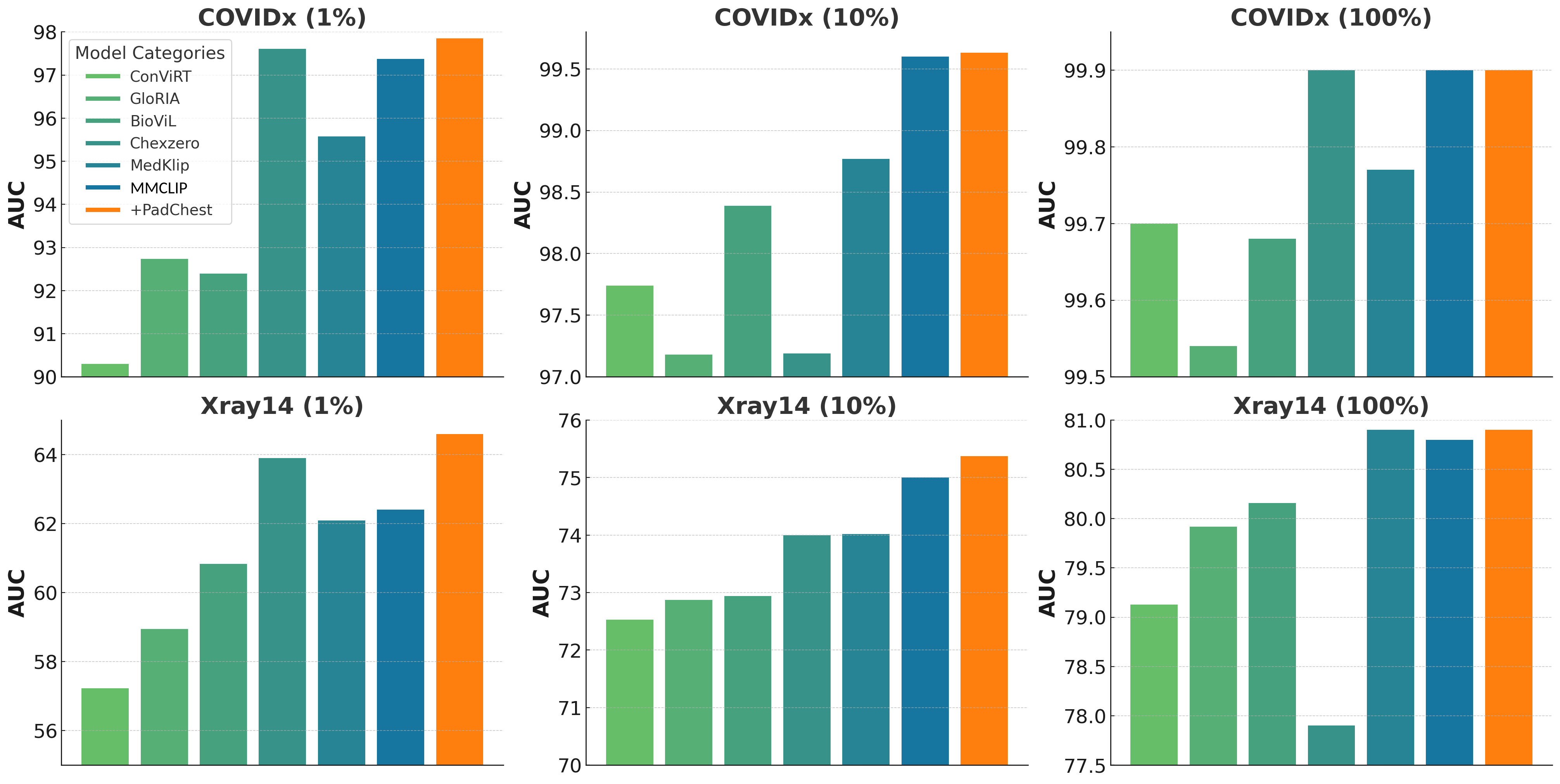} }

\caption{Performance comparison under different labeled data for fine-tuning. Unpaired data like PadChest can work with pair data to improve image encoder performance on the fine-tuning task.}  
\label{fig:image_5}  

\end{figure}

\subsection{Fine Tuning} 

Existing multi-modal frameworks are constrained as they require data that include both text and images, which limits the potential for the image encoder to leverage image-only data to improve its representation capabilities. To address this, we introduce a training approach designed to work effectively with unpaired, single-modal data. This new method allows the image encoder to better use image-only data, enhancing its representation power and overall performance. Our experiments involved two datasets, where we used 1\%, 10\%, and 100\% of the data for fine-tuning, aligning with the methods in previous works \cite{wu2023medklip}. As shown in Figure \ref{fig:image_5}, our model significantly outperforms existing models in terms of AUC scores across all datasets. Meanwhile, the performance of the model was further improved by incorporating image data from PadChest, demonstrating that compatibility with unpaired data is vital for representation learning.

\subsection{Ablation Study}

We noticed the original CheXpert validation set's limitations due to its small size and inconsistent performance tendency with the test set. To improve evaluation reliability, we crafted a new validation set with 2200 image-text pairs from existing datasets, which was used for all subsequent experiments, offering a more stable basis for performance assessment.

\subsubsection{Improvements from AttMIM} 

As shown in Table \ref{tab:AttMIM}, when we randomly mask 75\% of the features according to the setting of MIM in MAE \cite{he2022masked} and directly splice this module with CLIP and initialize it with the weights of MAE decoder, we can see that the model performance improves by 0.42\% of AUC score when we train it directly. However, CLIP and MIM rely on different training strategies, with the latter tending to require smaller learning rates and longer training epochs. To harmonize the two, we first train the image encoder via the MIM reconstruction task for 15,000 iterations, as warm Up process. With the Warm Up, the model further improves the AUC by 1.1\%. Meanwhile, we comparatively validate varying single strategies with mask ratios of 0.75, and the AttMIM strategy with the fusion of the three strategies. Experimental results show that three attention-based mask strategies can help the model understand key features better than random. As a fusion of the three strategies, AttMIM can more comprehensively improve related capabilities.

\begin{figure}[h]
    \centering
    \includegraphics[width=0.6\linewidth]{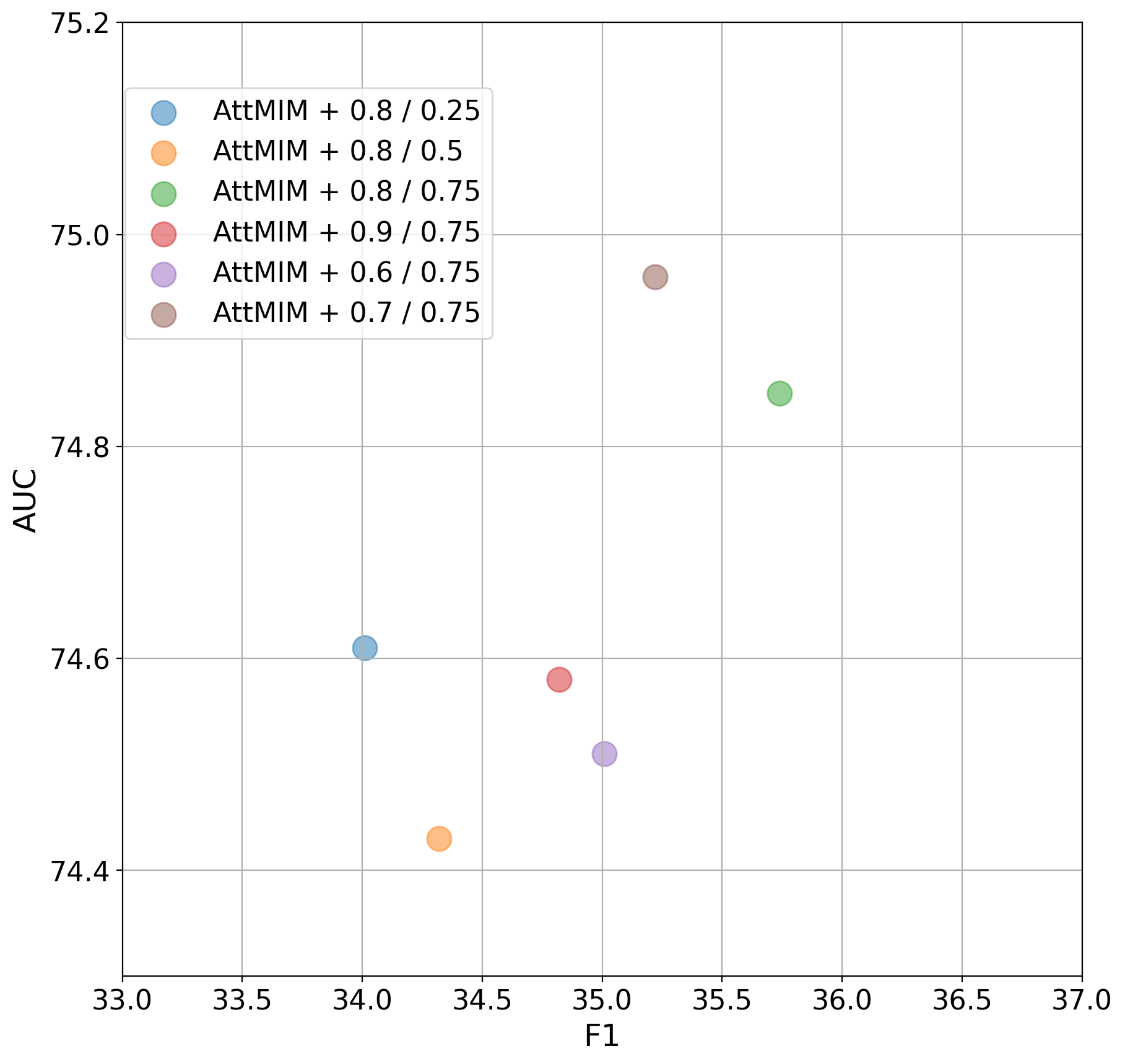}
    \caption{AUC and F1 under different mask ratios of AttMIM.}
    \label{fig:image_6}
\end{figure}

AttMIM has two adjustable parameters: the mask ratio of a single strategy and the final mask ratio after multi-strategy blending. As shown in Figure \ref{fig:image_6}, after comprehensively comparing the values of AUC and F1, we concluded that the best settings are 0.7 and 0.75, which means that all single strategies mask 70\% of the image features, and randomly select a subset after merging these masks, and finally mask 75\% of image features.

\begin{table}[t]
    \centering
    \caption{Zero-shot AUC on PadChest. 'Unseen 10' refers to the 10 most frequently unseen categories.}

    \resizebox{0.75\linewidth}{!}{%
    \begin{tabular}{llcc}
        \toprule
        Init & Mask Method & AUC & F1  \\
        \midrule
        no warm up & clip (Baseline)   &72.43 &34.24 \\
        no warm up & + MIM Random &72.85 &34.49  \\
        warm up &  + MIM Random    &73.98 &34.80  \\
        \midrule
        warm up &  + Image Report & 74.67 &34.71  \\
        warm up &  + Prompt-driven  & 74.91 &34.59  \\
        warm up &  + Image Self & 74.84 &34.70   \\
        \midrule
        warm up & + AttMIM & \textbf{74.96} & \textbf{35.22} \\
        \bottomrule
    \end{tabular}}
    \label{tab:AttMIM}

\end{table}

\begin{table}[t]
    \centering
    \caption{Performance of different text features masking strategies on the new CheXpert validation set.}

    \resizebox{0.75\linewidth}{!}{%
    \begin{tabular}{lcccc}
        \toprule
        Mask Method & Mask Ratio & AUC & F1 \\
        \midrule
        AttMIM  & - & 74.96  &35.22 \\
        + MLM Full   & 1.0 & 74.75 &34.79 \\
        + MLM Random  & 0.5 & 75.52 &34.51 \\
        \midrule
        + EntMLM & 0.5 & 74.98 & 34.49 \\
        + EntMLM & 0.3 & 75.44 & 34.88 \\
        + EntMLM & 0.1 & 74.67 & 34.74 \\
        + EntMLM & 0.2 & \textbf{75.78} & \textbf{35.02} \\
        \bottomrule
    \end{tabular}}
    \label{Ablation_MLM}

\end{table}

\subsubsection{Improvements from EntMLM} 

After splicing the MLM module with the existing structure, we first tried the Full mask and the Random mask based on the Causal Mask. The former did not make any modifications and essentially used MLM to complete the captioning task. The latter only randomly selected half of the Causal Mask to complete MLM tasks. As shown in Table \ref{Ablation_MLM}, we can find that a random mask 50\% is better than a Full mask. This may be because the task of predicting medical entities in medical texts is difficult. When there is a 50\% probability of seeing these entity words, it can help the model better learn the knowledge in the report. This highlights the importance of paying more attention to medical entity words. Since the medical text is a highly structured task focusing on the accuracy of medical entities rather than text fluency, the EntMLM we proposed mainly masks and reconstructs medical entity words. Experimental results confirmed this, highlighting the importance of mask medical entities.

\begin{figure}[h]
    \centering
    \includegraphics[width=0.6\linewidth]{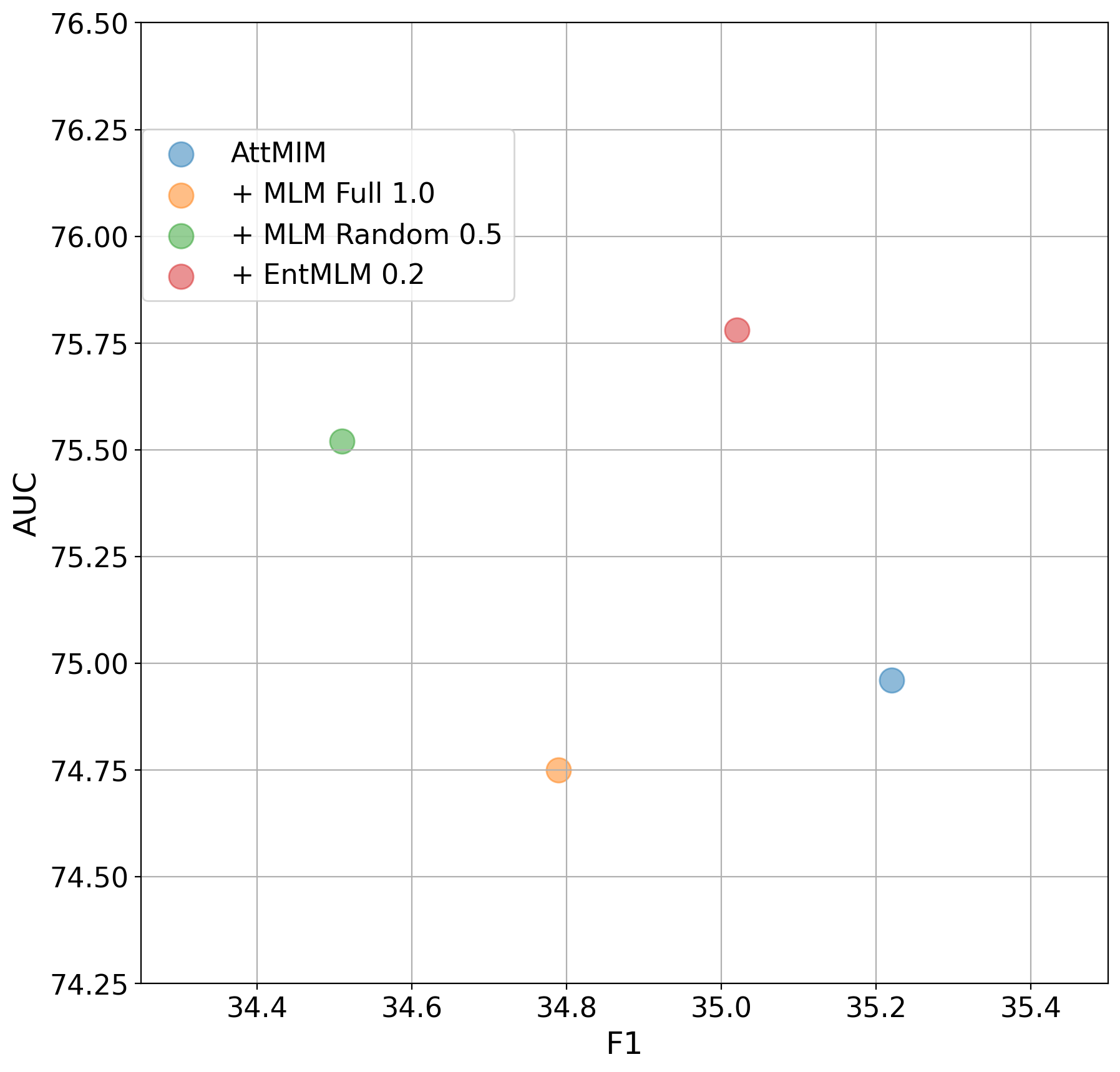}
    \caption{AUC and F1 under different mask ratios of EntMLM.}
    \label{fig:image_7}
\end{figure}

EntMLM has only one learnable parameter, unlike MLM where the mask ratio is a percentage of the sentence length. The ratio of EntMLM is the mask ratio for the medical entity words detected by NER. In Figure \ref{fig:image_7}, when the mask ratio is 0.2, the comprehensive indicators of AUC and F1 reach the best.

\section{Conclusion}

In this paper, we propose MMCLIP, an advanced medical VLP framework that aims to overcome major challenges in medical image and text analysis by introducing an innovative cross-modal attention masking modeling mechanism. By fusing AttMIM and EntMLM, MMCLIP can accurately capture key pathological features in medical data and improve the model's ability to understand complex medical scenes. In addition, we also use disease prompts to make the model compatible with unpaired data. Through pre-training on two large-scale medical datasets and validation on multiple downstream tasks, MMCLIP demonstrates its superior performance, not only achieving state-of-the-art results in the zero-shot setting but also in the fine-tuning setting. This result proves the practicality and effectiveness of MMCLIP in deep learning applications in the medical field.

\appendix
\vspace{0.5cm}
{\LARGE\textbf{Appendix}}
\vspace{-0.2cm}

\section{Pre-training Dataset}

\textbf{MIMIC-CXR \cite{johnson2019mimic}} encompasses more than 227k investigations with paired imagery and report data, originating from 65,379 individuals across varied scanning instances. Each investigation might contain one or two images (diverse scanning perspectives), summing up to a total of 377,110 images.

\noindent \textbf{PadChest \cite{bustos2020padchest}} is a comprehensive, chest x-ray dataset with 160k+ images from 67k patients, acquired at San Juan Hospital, Spain (2009-2017), inclusive of six varied positional views and rich associated patient/report data. Reports, categorized into 174 radiographic findings, 19 differential diagnoses, and 104 anatomical locations, utilize UMLS terminology and incorporate a hierarchical taxonomy. Limited by the fact that PadChest pairs are labelled in Spanish, this experiment only uses the image of this dataset.

\section{Datasets for Downstream Tasks}

\noindent \textbf{CheXpert \cite{irvin2019chexpert}}, is a large-scale dataset, containing 224,316 labelled chest X-rays from 65,240 patients. It covers 14 conditions including Fracture, Edema, Consolidation, Enlarged Cardiom, Cardiomegaly, Lung Lesion, Lung Opacity, Pneumonia, Atelectasis, Pneumothorax, Pleural Effusion, Pleural Other, Support Devices, and No Finding.  This dataset is involved in zero-shot and linear probing in our downstream tasks, both using the officially released test set with 500 images, which is consistent with the existing work \cite{wu2023medklip,xie2023medim}.

\noindent \textbf{COVIDx \cite{pavlova2022covid}} datasets is used for COVID-19 diagnosis. COVIDx has 29,986 images from 16,648 patients, used for the binary classification task of COVID symptoms, with a 70\%/20\%/10\% split for training, validation, and testing. This dataset is utilized for fine-tuning purposes, following the existing work \cite{wu2023medklip}.

\noindent \textbf{Pneumonia\cite{rsna-pneumonia-detection-challenge}} comprises over 260,000 frontal view chest X-ray images along with their corresponding pneumonia opacity masks, as well as support binary classification for pneumonia symptom, which were collected by the Radiological Society of North America. Following the settings of previous work\cite{wu2023medklip}, this dataset is divided into three parts: 60\% for training, 20\% for validation, and 20\% for testing.

\noindent \textbf{Xray14 \cite{wang2017chestx}} includes 112,120 frontal view X-ray images, each annotated with 14 disease labels derived from corresponding radiology reports. Following the methodology outlined in previous work\cite{wu2023medklip}, the dataset is divided into three subsets: 70\% of the data is allocated for training, 10\% for validation, and 20\% for testing. This division is carried out at the patient level to ensure that images from the same patient are not present across different subsets, thereby avoiding any overlap between the training, validation, and testing groups.

\noindent \textbf{PadChest testset} encompasses 193 different categories, which can be used to evaluate zero-shot classification performance for unseen classes. We first identify 20 categories that were present during the pretraining phase. Then, we pinpoint categories not seen during pretraining within the test set. Next, we rank these unseen categories based on their frequency and select groups of the top 10 and top 20 most frequent unseen categories, as well as a group including all 173 unseen categories. Performance tests for zero-shot classification are conducted separately on each of these groups.


\end{document}